# Discretization and fitting of nominal data for autonomous robots

Nuno Mendes, Pedro Neto, J. Norberto Pires and Altino Loureiro

*Department of Mechanical Engineering (CEMUC)-POLO II,*
*University of Coimbra,*
*3030-788 Coimbra, Portugal*
Phone: 00351 239 790 700
Fax: 00351 239 790 701
Email: nuno.mendes@dem.uc.pt

## Abstract

This paper presents methodologies to discretise nominal robots paths extracted from 3-D CAD drawings. Behind robot path discretization is the ability to have a robot adjusting the traversed paths so that the contact between robot tool and work-piece is properly maintained. In addition, a hybrid force/motion control system based on Fuzzy-PI control is proposed to adjust robot paths with external sensory feedback. All these capabilities allow to facilitate the robot programming process and to increase the robot's autonomy.

## Highlights

- Discretization and fitting of nominal robot paths extracted from 3-D CAD drawings;
- Discretization of robot paths allows to make adjustments in the pre-programmed paths;
- Implementation of an hybrid force/motion control system based on Fuzzy-PI control;
- The control system improves the contact conditions between the tool and work-piece;
- It was demonstrated that hybrid control improves significantly robot performance.

## 1 Introduction and Motivation

The process of robot off-line programming (OLP) has become increasingly popular in recent time, mainly inside Small and Medium-Sized Enterprises (SMEs). To this effect has contributed the required knowledge in robot programming which has been replaced by high widespread knowledge within enterprises. One way to use OLP is recurring to CAD drawings of the robotic cell in study (Kim, 2004; Vosniakos, & Chronopoulos, 2008; Liu, Bu, & Tan, 2010; Chen, &

Sheng, 2011). When CAD drawings are used to program a robot, the programming task can become easier, more intuitive and less monotonous. In addition, the knowledge required is high disseminated within enterprises structure because SMEs generally use CAD packages to design and develop their products. Moreover, while a robot is in the production phase the following working setup can be prepared off-line, thereby the setup time is reduced and the management of workers becomes easier.

A series of studies have been conducted using CAD as interface between robots and operators, for example, a review on CAD-based robot path planning for spray painting is presented by Chen, Fuhlbrigge, and Li (2009). Nagata, Kusumoto, Fujimoto, and Watanabe (2007) proposes a robotic sanding platform where the robot paths are generated by CAD/CAM software. An example of a novel process that benefits from the robots and CAD versatility is the so-called incremental forming process of metal sheets (Schaefer, & Schraft, 2005). The robot paths are achieved from a CAD model on the basis of specific material models. Prototype panels or customized car panels can be economically produced using this method. Feng-yun and Tian-sheng (2005) present a robot path generator for the polishing process, where cutter location data are generated from the postprocessor of a CAD system. A CAD-based robot programming system where is referred the use of a sequence of virtual robot tool models to define the robot paths is presented by Neto, Mendes, Araújo, Pires, and Moreira (2012). Nevertheless, the CAD-based robot programming systems have found some adversities. The major of them is calibration, i.e., the differences in alignment between the real environment and the virtual environment (CAD). These differences always exist and are almost impossible to determine because their origin is unpredictable. In order to deal with the uncertainty and inaccuracy of the robot working environment, the introduction of sensory-feedback in robotic systems has been studied and implemented (Mendes, Neto, Pires, & Moreira, 2010; Neto, Mendes, Pires, & Moreira, 2010). However, many of the robots are not able to easily incorporate sensory-feedback and some special care needs to be taken to ensure proper running of the system. One of the possible approaches is to discretize the path in small paths (Nagata, Kusumoto, Fujimoto, & Wtanabe, 2007). After that, some appropriate adjustments should be done on these small paths while the robot is performing them (Pires, Afonso, & Estrela, 2007). Several studies about robot path discretization have been proposed, such as (Nagata, Hase, Haga, Omoto, & Watanabe, 2007), which presents a CAD/CAM-based position/force controller. This study addresses a calculation method of robot orientation from cutter location data. Some algorithms for robot path discretization have been proposed to smooth robot paths (position and orientation) (Feng-yun, & Tian-sheng, 2005; Neto, Mendes, Araújo, Pires, & Moreira, 2012). Nielson (2004) proposes an algorithm for nonlinear smoothly interpolating orientation, which showed excellent results when used in Spline curves. Sheng, Xi, Song, and Chen (2001) present a method for robot path planning where nominal data are extracted from a CAD surface. A methodology for robot path planning from 3-D CAD models by means of the linear and circular discretization is presented by Berger and May (2005). Kuffner (2004) addresses some implementation issues and techniques in rigid body path planning using the *Lei group SE(3)*. Nagata (2005) presents a completely local algorithm for surface discretization. The main idea is the discretization of a curve segment supporting itself on the position and normal vectors at endpoints.

This paper aims to promote the OLP of industrial robots, in which nominal data (obtained for example from CAD drawings) are adapted to robot motion for industrial processes such as Friction Stir Welding (FSW) (Cook, Crawford, Clark, & Strauss, 2004; Soron, & Kalaykov, 2006; Fleming, Hendricks, Cook, & Strauss, 2009). The idea is to provide to users, with basic skills in CAD and robot programming, the tools to off-line generate reliable robot programs. However, as there are usually discrepancies between the virtual and real environment, the robot should be able to ensure recognition of their work environment and make appropriate adjustments (on-line) in the pre-programmed paths. In this way, we propose the use of sensory-feedback through the implementation of a hybrid force/motion control system. The function of this control system is to keep the contact between robot tool and work-piece. Nevertheless, even though not all commercial robots are ready to incorporate sensory-feedback in an easy way, most are prepared to make adjustments in the pre-programmed path, when the robot programs are being executed. Having this capability in mind, we propose the discretization of the nominal robot path in small sections and then make the adjustments that are desirable in these small sections (path adjustment). In this paper, some techniques of path discretization are presented, namely: linear paths, circular paths and curvilinear paths (Nagata patch). In addition, an interpolation technique is presented to interpolate end-effector orientation (Slerp). In order to perform the adjustments in the pre-programmed paths, a hybrid force/motion control system is presented, with special attention to the force control loop which is developed using Fuzzy reasoning and traditional Proportional Integrative control (PI).

The paper is further organized as followed. The second section focuses on the extraction of information from CAD. The third section presents some path discretization methods. The fourth section presents the hybrid force/motion control system used to produce adjustments in the pre-programmed path. The methods presented in the previews sections are evaluated through an experiment that is presented in section fifth. Finally, in the sixth section, conclusions are dawn and future work highlighted.

# 2 Acquisition and processing of nominal data from CAD

### 2.1 Case study

For this study, which is performed in the Cartesian space, nominal data are directly extracted from a commercial CAD package, Autodesk Inventor. Each CAD model consists in a robotic cell, points (representing robot paths also known as robot end-effector positions) and poses (representing orientations of the robot tool also known as robot end-effector orientations) Fig. 1. The data extracted from a CAD model are transformation matrices, consisting in rotation matrices and coordinates of points, both in relation to the origin of the CAD model of the cell. The information needed to program the robot will be extracted from the CAD environment by using an application programming interface (API) provided by Autodesk. This API allows the data transfer between the Autodesk Inventor and a Software Interface (which is the application that manages the entire process). Later, the information extracted from CAD is converted into robot code. A diagram with the procedure to extract nominal data from a CAD model and their conversion into a robot program is presented in Fig. 2.

In order to extract information from a CAD model, all the components belonging to the model (robot, work-pieces, conveyors, etc.) must be drawn and assembled like in the real cell in their real dimensions and shapes. Furthermore, the points must define the beginning and the end of the robot path. Nevertheless, it is the user that defines the type of path (linear, circular or curvilinear). Fig. 3 represents schematically a robot path. The arrows represent the robot tool orientation in each discretized position and the *P*'s represent robot positions.

## 2.2 Transformation between coordinate systems

In order to establish a match between the real environment and the virtual environment (CAD environment) some geometric references have to be provided. In other words, it is necessary to have all robot end-effector positions and orientations with respect to one or more reference frames (local coordinate systems) known a priori by the robot. These frames are made known to the robot through a calibration process. Generally, this is a simple and non-time consuming process where the user needs to define the frame(s) within the CAD environment by selecting three positions within the CAD drawing. Then, the user has to teach the real robot about that frame(s)' pose in the real scenario (off-line to on-line mapping). These three positions, which define a reference frame, have to follow some rules:

- The first position should be the origin of the frame;
- The second position should be a position in the Frame *x*-axis and non-coincident with the Frame origin;
- The third position should be a position in the Frame positive *xOy* quadrant and non-coincident with the *x*-axis of the Frame.

Aiming to convert a positional vector expressed in the coordinate system *{L}* into a positional vector expressed in the coordinate system *{M}*, the transformation matrix from *{L}* into *{M}*, ${}^{M}_{L}T$, must be calculated. Being *A*, *B* and *C* the first, second and third positions, respectively, defined in the three-dimensional space through coordinate system *{L}* and defining the vectors $\overrightarrow{AB}$ and $\overrightarrow{AC}$ from these positions as shown in the Fig. 4. We can calculate ${}^{L}_{M}T$ following the procedure below:

1. Calculate the vectors $\overrightarrow{AB}$ and $\overrightarrow{AC}$;
2. Calculate $\vec{E} = \overrightarrow{AB} \times \overrightarrow{AC}$;
3. Calculate $\vec{F} = \vec{E} \times \overrightarrow{AB}$;
4. Normalize the vectors $\overrightarrow{AB}$, $\vec{E}$ and $\vec{F}$.

Being $\vec{ab}$, $\vec{e}$ and $\vec{f}$ the normalized vectors of $\vec{AB}$, $\vec{E}$ and $\vec{F}$, respectively, the rotation matrix from *{L}* into *{M}* is defined as $^L_M\mathbf{R} = \begin{bmatrix} \vec{ab} & \vec{f} & \vec{e} \end{bmatrix}$. The transformation matrix can be represented by:

$$^L_M\mathbf{T} = \begin{bmatrix} \vec{ab} & \vec{f} & \vec{e} & A \\ 0 & 0 & 0 & 1 \end{bmatrix} \quad (1)$$

As our goal is to transform coordinate vectors expressed in coordinate system *{L}* into the new defined coordinate system *{M}*, the transformation matrix appears:

$$^M_L\mathbf{T} = \begin{bmatrix} ^L_M\mathbf{R}^T & -^L_M\mathbf{R}^T \cdot A \\ 0 & 1 \end{bmatrix} \quad (2)$$

To transform a robot position *P* expressed in a coordinate system *{L}* into a coordinate system *{M}* we have:

$$\vec{^M\mathbf{P}} = ^M_L\mathbf{T} \cdot \vec{^L\mathbf{P}} \quad (3)$$

## 3 Information processing and path discretization

When we want to discretize a robot path, two important factors should be taken in consideration: the type of movement and the orientation assumed by the robot when it performs such movement. In this study, it is assumed that the variation of orientation is proportional to the variation of position. However, this assumption cannot be true if the robot path is a curve with non-constant curvature (irregular arc segment). Such situation can lead to the existence of some small errors.

The paths analysed in this work were linear, circular and curvilinear with one concavity, which are the most used in common robotic applications. A method of orientation discretization is also presented.

### 3.1 Linear discretization

If the path to be followed by the robot end-effector is a linear path, the discretization method should be as follows. Considering two robot end-effector positions $P_A$ and $P_B$, if we want to move from $P_A$ to $P_B$ by a linear movement, as shown in the Fig. 5, the first thing we need to do is to define how long the distance between the discretized positions, *k*, should be. Consequently, the number of intermediate positions, *n*, should be calculated by:

$$n = \frac{\|\vec{P_A P_B}\|}{k} \quad (4)$$

Now, the intermediate positions, $P_j$, can be calculated:

$$P_j = P_{j-1} + k \cdot \overrightarrow{p_A p_B} \quad , \quad j \in \mathbb{N} \land j \subset [1, n-1]$$
$$P_0 = P_A \tag{5}$$
$$P_n = P_B$$

Where $\overrightarrow{p_A p_B}$ is the normalized vector of $\left\| \overrightarrow{P_A P_B} \right\|$.

Linear discretization demonstrated to be a good method to apply in any situation of linear path (Cartesian space) because it is computational efficient (in terms of computation time) and it is an exact method (it does not present any residual error). Furthermore, it is simple and easy to implement because only a few calculations are involved. In this way, linear discretization is suitable to be incorporated into the robot programming code. Nevertheless, there is a curiosity that concerns to the "last movement". This one can be smaller than the other movements, Fig. 5. This phenomenon occurs because we define the distance between positions, called *k*, and not the number of discretized positions, *n*. However, this is not a problem, it is a difference between movements' length, which does not put the task in risk.

### 3.2 Circular discretization

In order to have a versatile circular discretization method, three positions belonging to the nominal path should be provided ($P_1$, $P_2$ and $P_3$). The first one should be the starting position, the second one should be an intermediate position and finally the third position should be the finishing position, as represented in Fig. 6. Through these positions the centre of the arc $P_0=(x_0,y_0,z_0)$ and the radius *r* can be calculated by:

$$\begin{cases} (x_1 - x_0)^2 + (y_1 - y_0)^2 + (z_1 - z_0)^2 = r^2 \\ (x_2 - x_0)^2 + (y_2 - y_0)^2 + (z_2 - z_0)^2 = r^2 \\ (x_3 - x_0)^2 + (y_3 - y_0)^2 + (z_3 - z_0)^2 = r^2 \\ \begin{vmatrix} x_1 - x_0 & y_1 - y_0 & z_1 - z_0 \\ x_2 - x_0 & y_2 - y_0 & z_2 - z_0 \\ x_3 - x_0 & y_3 - y_0 & z_3 - z_0 \end{vmatrix} = 0 \end{cases} \tag{6}$$

The generation of the discretized positions using circular discretization is processed between $P_1$ and $P_3$. In the first step, a local coordinate system is defined, in which its origin is $P_0$ and the direction along the *x*-axis is the same direction as $\overrightarrow{P_0 P_1}$. The direction of the *y*-axis is defined by $P_2$, which is located in the first quadrant of the *xOy* plane as shown in section 2.2. In the second step it is computed the angle $\theta$ formed between $\overrightarrow{P_0 P_1}$ and $\overrightarrow{P_0 P_3}$, (7), $\theta_{inc}$, (8) (where $l_0$ is the increment of distance travelled in the arc) and the number of generated positions *n*, (9).

$$\theta = \cos^{-1}\left( \frac{\overrightarrow{P_0 P_1} \cdot \overrightarrow{P_0 P_3}}{\left\| \overrightarrow{P_0 P_1} \right\| \cdot \left\| \overrightarrow{P_0 P_3} \right\|} \right) \tag{7}$$

$$\theta_{inc} = \frac{l_0}{r} \tag{8}$$

$$n = \frac{\theta}{\theta_{inc}} \qquad (9)$$

In the third step the discretized positions are generated by:

$$\begin{cases} x_i = r \cdot \cos(\theta_{inc} \cdot i) \\ y_i = r \cdot \sin(\theta_{inc} \cdot i) \\ z_i = 0 \end{cases}, \quad i \in \aleph \ \wedge \ i \subset [0, n] \qquad (10)$$

In the fourth and last step, the generated positions are transformed from the local coordinate system to the global coordinate system as presented in section 2.2.

This circular discretization method presents a very similar behaviour to the method presented in section 3.1. Once again, it is an exact method and thus there is no theoretical error. Furthermore, the "last movement" is smaller in some situations for the same reasons that were pointed out in the previous method. However, the task is not put in jeopardy. Fig. 6 presents the discretized positions through the circular discretization method.

### 3.3 Curvilinear discretization (Nagata patch)

In order to obtain an efficient computational method for curvilinear discretization we decided to use a numerical method which was recently proposed by Nagata (2005) and has already been applied successfully to diverse engineering problems (Neto, Oliveira, Alves, & Menezes, 2010; Neto, Oliveira, Menezes, & Alves, 2012; Boschiroli, Fünfzig, Romani, & Albrecht, 2011; Sekine, & Obikawa, 2010). The main idea behind this parametric surface description (subsequently named Nagata patch) is the quadratic discretization of a curved segment from the position and normal vectors at the end points. The method here presented was initially developed for high-precision milling (Lin, Watanabe, Morita, Uehara, & Ohmori, 2007). However, this study aims to show that Nagata patch can be applied to robot path definition.

This method starts with the calculation of the rotation centre $P_0$, (6). For this purpose, three positions belonging to the curve have to be provided ($P_1$, $P_2$, $P_3$), Fig. 7. After that, the discretization is processed, in a first phase among the positions $P_0$, $P_1$ and $P_2$. In a second phase the second part of curve is discretized among the positions $P_0$, $P_2$ and $P_3$. Owing to the repetition of the method, in this document is only presented in detail the first phase of discretization. Fig. 8 shows the centre of curvature, $P_0$, the initial position, $P_1$ and the final position, $P_2$. The directional vectors $\vec{n_1}$ and $\vec{n_2}$ are then calculated (11) and (12), as well as the shortest vector connecting $P_1$ to $P_2$, $\vec{d}$ (13).

$$\vec{n_1} = \frac{P_1 - P_0}{\|P_1 - P_0\|} \qquad (11)$$

$$\vec{n_2} = \frac{P_2 - P_0}{\|P_2 - P_0\|} \qquad (12)$$

$$\vec{d} = P_2 - P_1 \qquad (13)$$

Before applying the curvilinear discretization method using Nagata patch (16), $\vec{c}$ (15) and $a$ (14) have to be calculated. The angle among the centre of curvature and the initial and final positions, $\alpha$, is limited to a value lower than 180°, Fig. 8.

$$a = \vec{n_1} \cdot \vec{n_2} = \cos(\alpha) \quad (14)$$

$$\vec{c} = \begin{cases} \dfrac{1}{1-a^2} \cdot [\vec{n_1} \; \vec{n_2}] \cdot \begin{bmatrix} 1 & -a \\ -a & 1 \end{bmatrix} \cdot \begin{Bmatrix} \vec{n_1}^T \cdot \vec{d} \\ -\vec{n_2}^T \cdot \vec{d} \end{Bmatrix}, a \neq \pm 1 \\ \vec{0}, a = \pm 1 \end{cases} \quad (15)$$

$$P(\eta) = P_0 + (\vec{d} - \vec{c}) \cdot \eta + \vec{c} \cdot \eta^2 \;,\; \eta \in \Re \;\wedge\; \eta \subset [0, 1] \quad (16)$$

In order to estimate the parameter $\eta$ we have to determine the number of movements (position increments), $n$, by calculating the total curve length to be travelled. To this end, we do an approximation of the curve to a perfect arc (17), (18), (19) and (20).

$$\alpha = \cos^{-1}(a) \quad (17)$$

$$\alpha_0 = \frac{l_0}{r} \quad (18)$$

$$n = \frac{\alpha}{\alpha_0} \quad (19)$$

$$\eta_i = \frac{1}{n} \quad (20)$$

Where $\alpha$ is the total curve angle to be travelled, $\alpha_0$ is the curve angle to be travelled in each increment, $l_0$ is the curve length to be travelled in each increment and $\eta_i$ is the increment of the $\eta$ in each increment.

      The Nagata patch proved to be a very good method for representing curves with a concavity. To illustrate the power of the Nagata method, it was used to represent two well-defined curves, a parabolic curve and an arc, using only three positions for each one. The results can be seen in Fig. 9 and Fig. 10, and Table 1 and Table 2. Since both curves are symmetric, in the tables are only shown the results for the first halves of the curves. The analysis allowed concluding that there are small differences in the generated positions and the well-defined curves. In order to generalize these differences, they are presented in Table 1 and Table 2 as a function of the radius of the curvature because of the influence of this parameter in the magnitude of the differences. Another parameter that influences the difference between the curves is the rotated angle $\alpha$, i.e. higher angles (larger distances between positions) lead to major differences. The maximum difference occurs in the half angle of the total angle, i.e., for an angle of 90°, the maximum error occurs at an angle of 45°. Although the formulation presented above is only defined for a specific

kind of curvature orientation, it is possible to increase or decrease the curvature of the curve by defining the vectors $\vec{n_1}$ and $\vec{n_2}$ in a different way.

The major advantages of the Nagata patch are:
- Computationally efficient;
- It works over a local coordinate system;
- The method uses only a variable and this one is dimensionless;
- It is compact, fast and easy to implement.

The errors obtained in a path discretization through this method are acceptable when used in robotic tasks that do not require very strict accuracy, for example in FSW. In spite of the good results presented by this method in the representation of curves with a concavity, Nagata patch should not only be seen as a method of representation of paths which were previously generated by other methods. The Nagata patch is a different method for path generation and it should be faced as well, an advanced method that enables the generation of a given type of path.

### 3.4 Orientation interpolation (Slerp)

A quaternion interpolation algorithm (spherical linear interpolation – Slerp) to interpolate smoothly a sequence of end-effector orientations is here presented (Fig. 11). Given two known unit quaternion, $Q_0$ and $Q_n$, with parameter *k* moving from *1* to *n-1*, the interpolated end-effector orientation $Q_k$ can be computed:

$$Q_k = \frac{\sin\left(\left(1-\frac{k-1}{n-1}\right)\cdot\theta\right)}{\sin\theta}\cdot Q_0 + \frac{\sin\left(\frac{k-1}{n-1}\cdot\theta\right)}{\sin\theta}\cdot Q_n \ , \ k\in\aleph \ \wedge \ k\subset[1 \ \ n-1]$$

(21)

Where:

$$\theta = \cos^{-1}(Q_0 \cdot Q_n) \quad (22)$$

The method used to interpolate the orientation is very efficient and reliable for this kind of robotic application. If we have two robot positions which belonging to an arc and the orientation of these positions is perfectly perpendicular to the arc, the orientation of these discretized positions will be perfectly perpendicular to that arc. This means that the orientation increment is linear. Nevertheless, when this method is used for curved paths the change of orientation along the path should not be linear, in some cases, because the curvature can be more accentuated in the beginning or in the ending of the path. In order to solve this issue another methods should be studied in the future as the v-Spline orientation method (Nielson, 2004). In Fig. 12 is presented some orientation interpolation of robot end-effector path positions for different kinds of robot paths.

# 4 Hybrid force/motion control system

Since there may be small differences between pre-programmed robot paths, through off-line programming, and real environments, we propose the use of a hybrid force/motion control system to attenuate or even avoid the influence of these differences. This strategy is based on sensory-feedback in which a force/torque sensor feeds the controller. The controller consists in an external force control loop which in turn feeds an internal position control loop as in Fig. 13. The external control loop receives force and torque data about the contact between robot and its environment and sends robot displacement data (adjustments) to the internal control loop, as can be seen in Fig. 14. The Fuzzy reasoning and the Proportional Integrative Control (PI) were used to obtain a powerful and simple control framework.

The characteristics of this controller should be simplicity, flexibility and fast to compute. A null steady state error is pretended to be achieved and in this way a PI-type control was chosen. The association of the two powerful mathematical tools, Fuzzy Logic and PI control, allow to use the intuitive knowledge from users and the good performance of the PI control already so well known. Thus, a Fuzzy logic controller type Mamdani (Mamdani, 1974; Bingül, & Karahan, 2011) based on the traditional PI controller was implemented. This kind of controller is easy to implement, since it does not need a precise mathematical model of the system. The controller is implemented only using linguistic variables to model the intuitive knowledge of the operator. The implementation of the Fuzzy controller should respect the following order:

- Definition of input and output variables.
- Fuzzification.
- Definition of a group of rules to model the application in study (knowledge base).
- Design of the computational unit that accesses the Fuzzy rules.
- Defuzzification.

The force control loop collects inputs from the force/torque sensor and processes these inputs according to the rules specified in the Fuzzy logic memberships (section 4.2). The outputs $\Delta \mathbf{u}$ are presented as displacement of the end-effector at which the robotic arm should be moved to obtain the desired contact range of forces and torques.

## 4.1. Fuzzy-PI controller

In our system, the controller input variables are the force/torque error **e** and change of the error **de**:

$$\mathbf{e}_k = \mathbf{f}_{\mathbf{d}_k} - \mathbf{f}_{\mathbf{e}_k} \quad (23)$$

$$\mathbf{de}_k = \mathbf{e}_k - \mathbf{e}_{k-1} \quad (24)$$

Where, $\mathbf{f_e}$ is the actual wrench and $\mathbf{f_d}$ is the desired wrench (set-point). The controller output is the position accommodation for the robot (established in terms of robot displacements).

From the discrete version of traditional PI controller:

$$u_k = u_{k-1} + \Delta u_k \quad (25)$$

$$\Delta u_k = K_P de_k + K_I e_k \quad (26)$$

Where, **u** is the robot displacement; and $K_P$ and $K_I$ are coefficient constants. If **e** and **de** are Fuzzy variables, (8) and (9) become a Fuzzy control algorithm. A practical implementation of the proposed Fuzzy-PI concept is simplified in Fig. 14. Finally, the centre of area method was selected for defuzzify the output Fuzzy set inferred by the controller:

$$\Delta U = \frac{\sum_{i=1}^{n} \mu_i \cdot \Delta U_i}{\sum_{i=1}^{n} \mu_i} \quad (26)$$

Where, $\mu_i$ is the membership function, which takes values in the interval [0, 1].

A decision maker $S \in \Re^{n \times n}$ establishes the axis direction we want to control with force/motion control action. When all the axes are controlled with force and motion actions, the **S** matrix becomes the identity matrix.

### 4.2. Knowledge base

The knowledge base of a Fuzzy logic controller is composed of two components, namely, a database and a Fuzzy control rule base. Each control variable should be normalized into seven linguistic labels {PL , PM , PS , ZR , NS , NM , NL}. The grade of each label is described by a Fuzzy set. The function that relates the grade and the variable is called the membership function (Fig. 15). The well-known PI-like Fuzzy rule base suggested by MacVicar-Whelan (1977) is used in this paper (Table 3). It allows fast working convergence without significant oscillations and prevents overshoots and undershoots.

### 4.3. Tuning strategy

Fuzzy logic design is involved with two important stages: knowledge base design and tuning. However, at present there is no systematic procedure to do the tuning. The control rules are normally extracted from practical experience, which may make the result focused in a specific application. The objective of tuning is to select the proper combination of all control parameters so that the resulting closed-loop response best meets the desired design criteria.

In order to adapt the system to different contact conditions, the scaling factors should be tuned. The controller should be adjusted with characteristics representing the working environment to be controlled. These adjustments can be made through the scaling factors, usually applied in any PI controller, namely $K_P$, $K_I$ and $K_X$. Lin, and Huang (1998) proposes an adjustment where the scaling factors are dynamic and thus they have been adjusted along the task. Similar strategies are proposed in references (Yu, 2009; Mudi, & Pal, 2000; Chen, Tung, Tsai, & Fan, 2009). The utilization of different tables of rules accordingly the task to be performed and the materials involved are presented by Pires, Godinho, and Araújo (2004). Dynamic membership functions are proposed by Woo, Chung, and Lin (2000). In this study, the scaling factors are set to appropriate constant values, achieved by the method of trial and error.

## 5. Experiments

The effectiveness of the proposed approach was evaluated in an experiment, which was repeated 20 times. In this experiment, the robot end-effector should follow a specific robot path keeping the contact between robot end-effector and surface and avoiding excessive contact forces. The robot path was pre-programmed by a CAD drawing, Fig. 1, as presented in section 2 as well as intended calibrations were done. The real path followed by the robot end-effector is presented in Fig. 16. Note that the surface is highly irregular, which makes this experiment very challenging for any control system. In order to verify the efficiency of the proposed control system two tests of the experiment were checked. In a first test the hybrid control system was implemented to control the normal force to the contact surface. In a second test the robot end-effector execute the pre-programmed path without force control loop, i.e. the robot end-effector path is not adjusted. In both tests the normal force to the contact surface was registered and compared.

The Fig. 17 and Fig. 18 show the results for the two tests of the experiment, all experimental results presented similar behaviours to these ones. By the analysis of Fig. 17, without make adjustments in the pre-programmed path, we can conclude that the contact between the robot tool and surface is lost several times. Furthermore, the contact force reaches high values as can be seen in the same figure. The experiment failed (excessive contact force) when the discrete time reached the value 323, in which there was the need to stop the experience due to the high contact force values. After that, the experiment only could be restarted for the discrete time 344 (where the safety conditions were satisfied). Notice that even at the point where the experiment was restarted the value of the contact force is very high. The discrete time is directly associated with the positions that compose the robot end-effector path.

In relation to the Fig. 18, which represents the other test, using the force control loop, it can be seen that the contact force is more carefully controlled. Although the contact is lost and the contact force is high in some cases, it is clear the tendency of the contact force to converge to a set-point value. By the comparison between Fig. 17 and Fig. 18, the efficiency of the force control loop is evident. With the use of this system the high contact force values are lower and the contact is lost less times. In the analysis of these data it is necessary to take into account that the stiffness of the system is high, the tool is made of steel and the work-piece is made of wood, hence the contact force vary significantly from point to point. These figures do not allow to have a perspective of how far the two interacting parts are one from another and this point can lead to wrong judgments. When the contact is lost, it is difficult to find a balance between to get in contact quickly and to avoid large overshoots (excessive contact forces). Hence we have chosen to work in the security side, this way large deviations take long times correction.

## 6. Conclusions and future work

In order to take advantages of CAD-based OLP for robotic applications that require a considerable high degree of accuracy, the self-recognition of the robot work environment (by means of sensory-feedback) should be achieved. However, not all commercial industrial robots allow easy sensory-feedback integration. Our proposal is to discretize all nominal robot end-effector paths, extracted from a 3-D CAD drawing, where it is intended to make adjustments. Then the discretized robot end-effector paths are on-line adjusted by the use of a hybrid force/motion control system. With this idea in mind, some discretization methods were presented, namely: discretization method of linear path, circular path, curvilinear path (Nagata patch) and a method of orientation interpolation (SLERP). The two firsts methods (linear and circular discretization) are exact methods so that they have no errors, this analysis was done taking into account the robot Cartesian space. On the other hand, the Nagata patch is a numerical method and presents small errors, which do not put the robot task at risk. With regard to the orientation interpolation method, it interpolates the orientation proportionally to the number of points generated, being constant the increment of orientation. This works very well to linear and circular paths but when used with curved paths, where the linearity is not verified, some slight disproportionality can occur. In order to solve this issue another methods should be studied in the future between them the v-Spline orientation method.

In order to produce adjustments in the pre-programmed robot paths a hybrid force/motion control system was proposed, a system based on an external force control loop and an internal position control loop. In this study it is only detailed the external control loop, which consists in a Fuzzy-PI controller that receives force/torque data and sends position displacements to the robot end-effector. The effectiveness of this system was tested in an experiment divided in two tests, where in one of the tests a path was traversed by a robot using the hybrid control system and in the other test the same experiment was repeated using a position control system, i.e. using the same hybrid control system without the external force control loop. The benefit of using the proposed system is clear, being the great advantages the reducing number of points of loss of contact and the diminution of the excessive contact forces as well as there was a clear tendency in the contact force to converge to a set-point. The convergence to the set-point can be improved in order to maintain a more stable contact force, with less fluctuation, and more accuracy. For this purpose, more labels and more rules should be incorporated in the Fuzzy-PI controller. General speaking, the presented approach overcame the challenges, thus, OLP becomes more appealing to be used and is a step more to achieve the concept of autonomous robots.

# Research highlights

- Discretization and fitting of nominal robot paths extracted from 3-D CAD drawings;
- Discretization of robot paths allows to make adjustments in the pre-programmed paths;
- Implementation of an hybrid force/motion control system based on Fuzzy-PI control;
- The control system improves the contact conditions between the tool and work-piece;
- It was demonstrated that hybrid control improves significantly robot performance.

**Table 1**

Table 1 Error present in a Nagata patch test.

| | First section ($\alpha \approx 33°$) | | | | | | | | | | |
|---|---|---|---|---|---|---|---|---|---|---|---|
| % Path per section | 0 | 10,3 | 20,5 | 30,7 | 40,7 | 50,7 | 60,7 | 70,6 | 80,4 | 90,2 | 100 |
| Error/Radius (%) | 0 | -3 | -5,1 | -6,4 | -6,8 | -6,6 | -5,8 | -4,6 | -3,1 | -1,5 | 0 |
| | Second section ($\alpha \approx 21°$) | | | | | | | | | | |
| % Path per section | 0 | -3,4 | -6,6 | -9,8 | -12,8 | -15,8 | 100 | | | | |
| Error/Radius (%) | 0 | 1,2 | 2,1 | 2,5 | 2,4 | 1,6 | 0 | | | | |

**Table 2**

Table 1 Error presented in a parabolic path generated by Nagata patch.

|  | First section | | | | | | | | | | |
|---|---|---|---|---|---|---|---|---|---|---|---|
| % Path per section | 0.0 | 11.5 | 23.1 | 38.5 | 42.3 | 46.2 | 61.5 | 69.2 | 76.9 | 88.5 | 100 |
| Error/Radius (%) | 0.0 | -6.1 | -10.2 | -12.4 | -12.5 | -12.4 | -10.4 | -8.6 | -6.6 | -3.2 | 0.0 |
|  | Second section | | | | | | | | | | |
| % Path per section | 0.0 | 8.3 | 16.7 | 25.0 | 33.3 | 50.0 | 58.3 | 66.6 | 83.3 | 91.7 | 100 |
| Error/Radius (%) | 0.0 | 0.9 | 1.7 | 2.4 | 2.9 | 3.5 | 3.5 | 3.3 | 2.2 | 1.2 | 0.0 |

**Table 3**

**Table 1** Representation of the Rule Base.

| de \ e | NL | NM | NS | ZR | PS | PM | PL |
|---|---|---|---|---|---|---|---|
| **PL** | nl | nm | ns | zr | pm | pl | pl |
| **PM** | nl | nl | nm | zr | pm | pl | pl |
| **PS** | nl | nl | ns | zr | ps | pl | pl |
| **ZR** | nl | nm | ns | zr | ps | pm | pl |
| **NS** | nl | nl | ns | zr | ps | pl | pl |
| **NM** | nl | nl | nm | zr | pm | pl | pl |
| **NL** | nl | nl | nm | zr | ps | pm | pl |

**Figure 1**

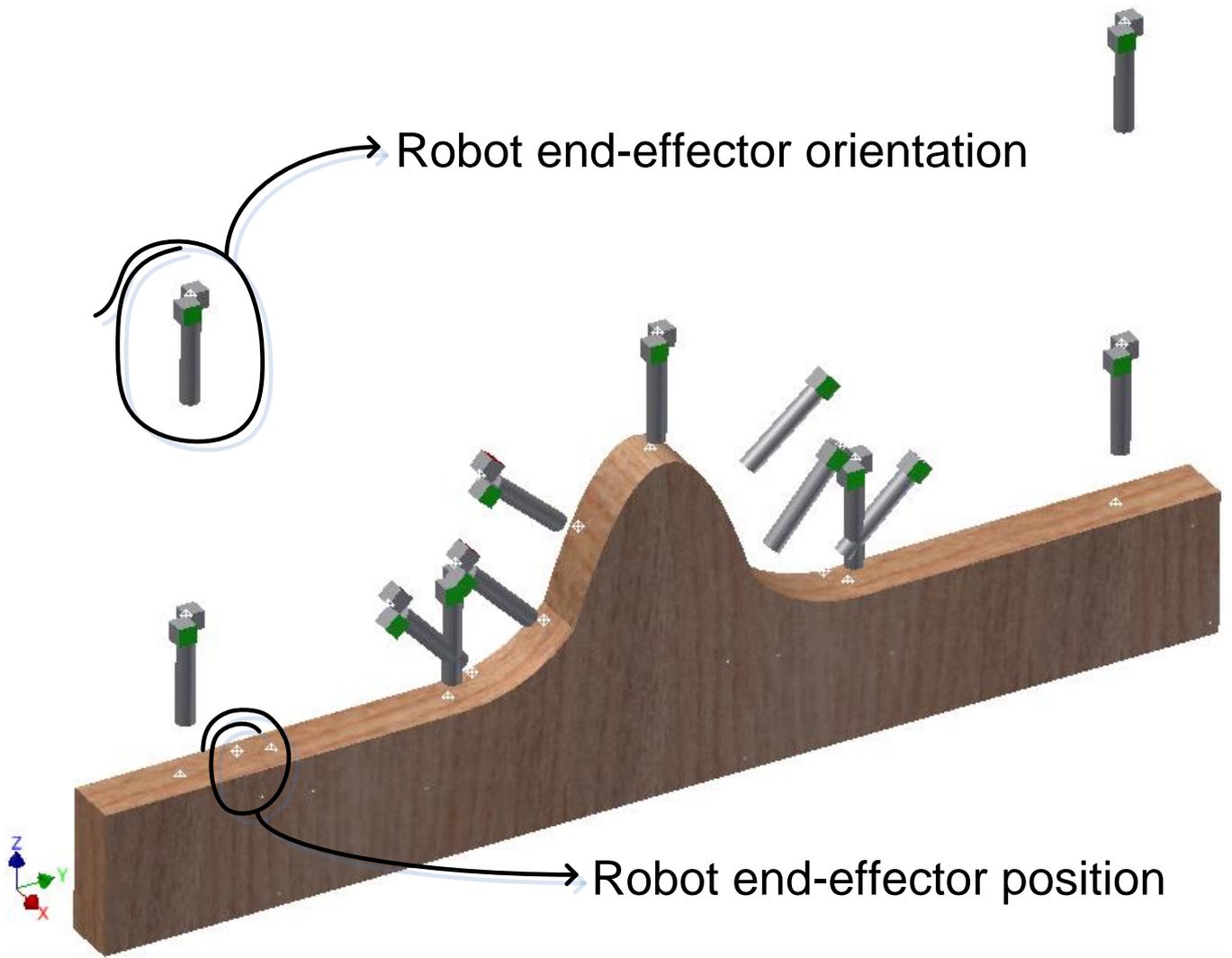

**Fig. 1** A CAD model.

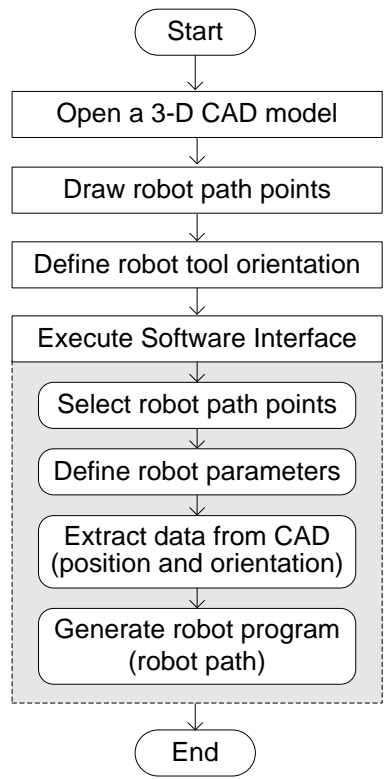

**Fig. 1** Procedure to extract nominal data and generate robot programs from CAD.

**Figure 3**

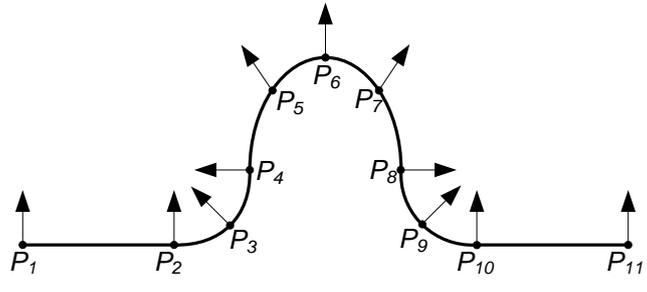

**Fig. 1** Schematic representation of a 3D CAD path to extract nominal data.

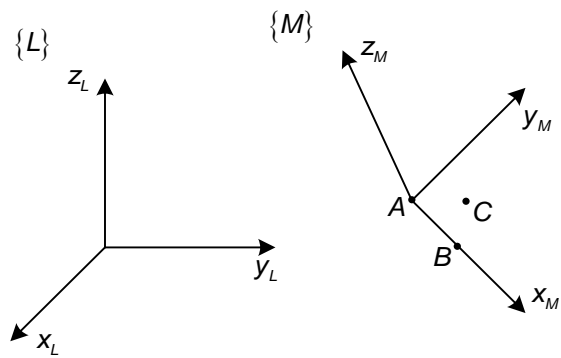

**Fig. 1** Defining a frame.

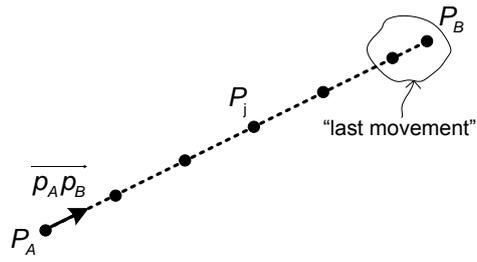

**Fig. 1** Representation of the linear discretization method.

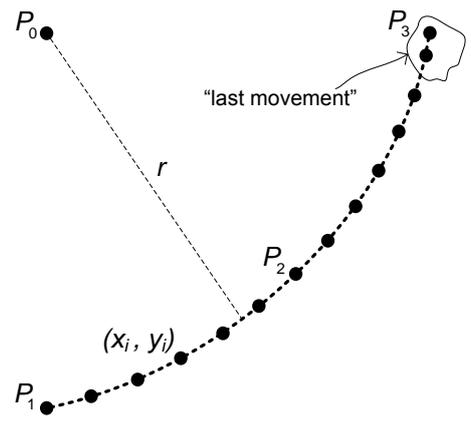

**Fig. 1** Nominal data for circular discretization.

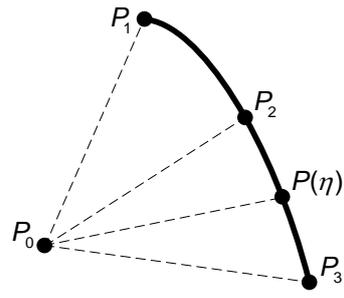

**Fig. 7** Nominal data for Nagata patch.

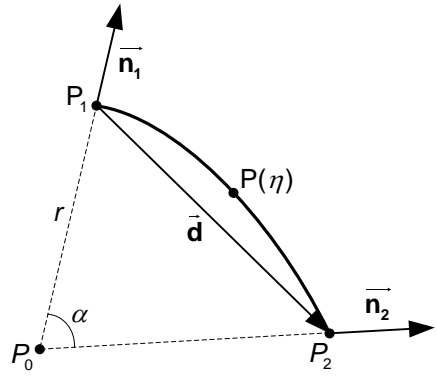

**Fig. 8** Representation of the Nagata Patch.

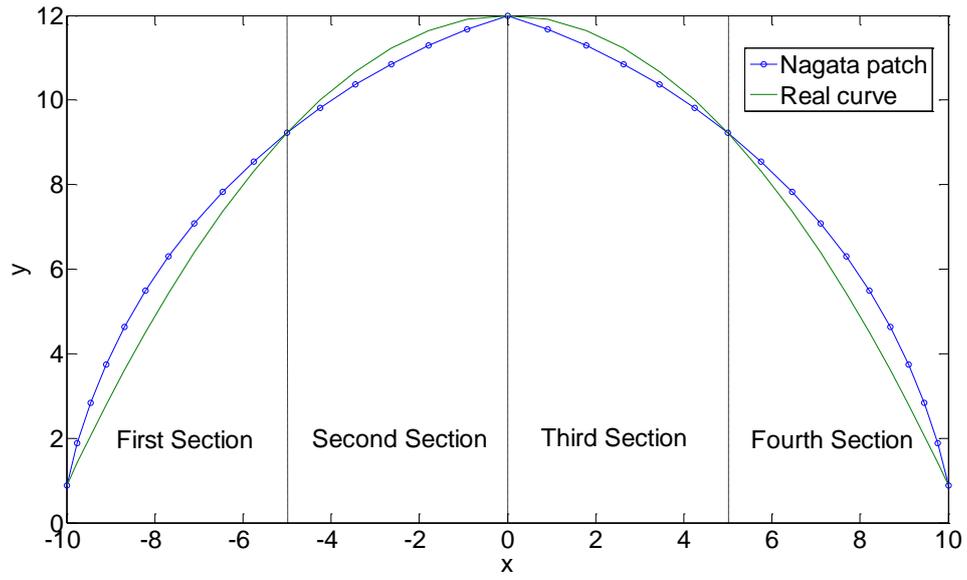

**Fig. 9** Error presented in a circular path generated by Nagata patch.

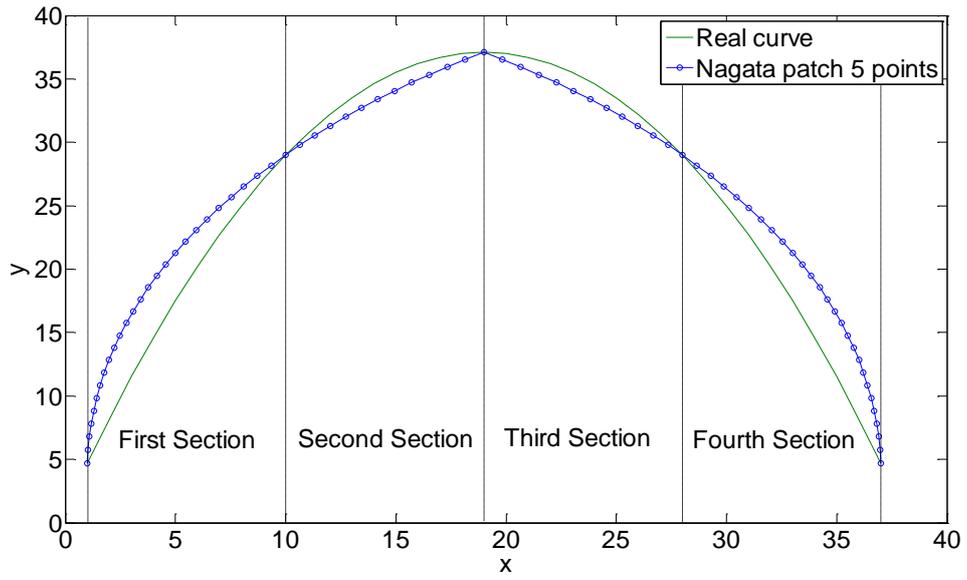

**Fig. 1** Error presented in a curvilinear path generated by Nagata patch.

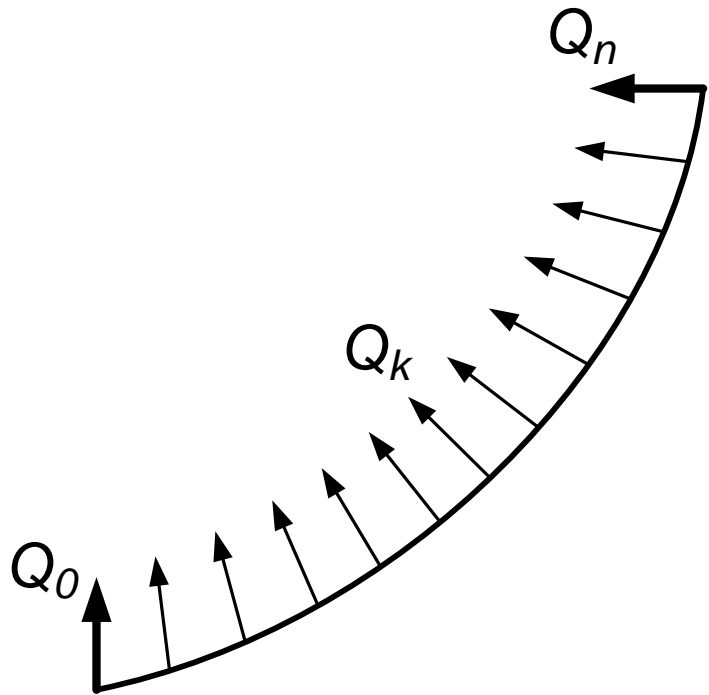

**Fig. 11** Orientation interpolation.

**Figure 12**

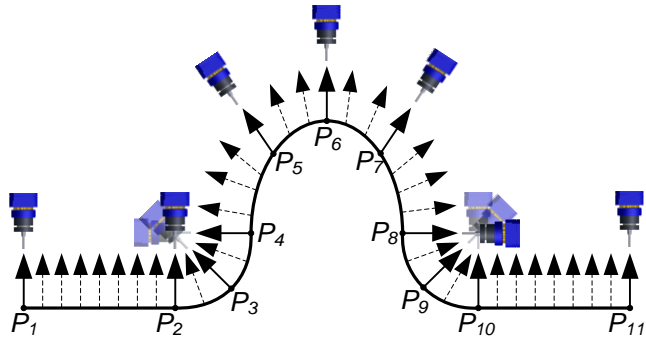

**Fig. 12** Orientation interpolation of robot paths.

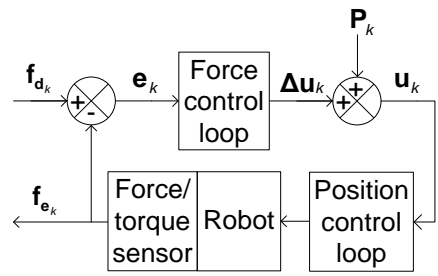

**Fig. 13** Hybrid force/motion control system.

**Figure 14**

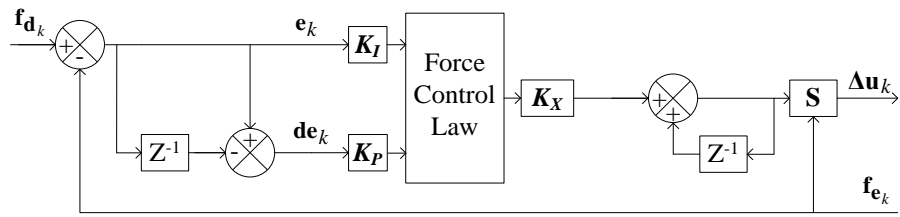

**Fig. 14** External force control loop.

**Figure 15**

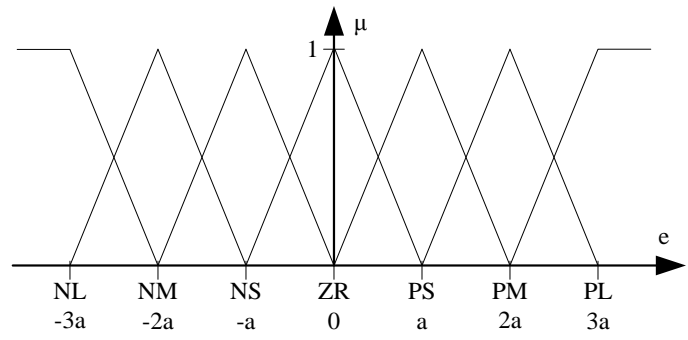

**Fig. 15** Membership functions for the input variables.

**Figure 16**

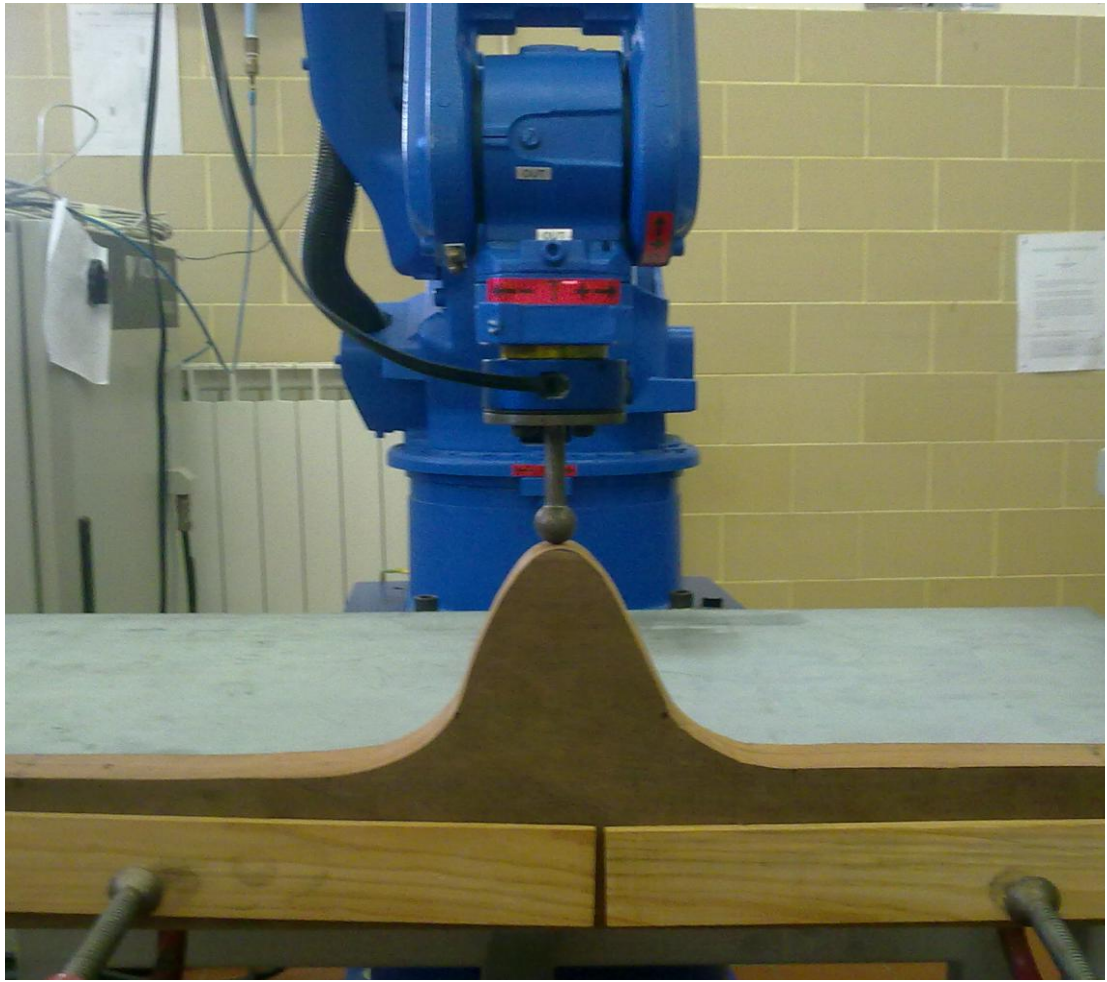

**Fig. 16** Experiment layout.

**Figure 17**

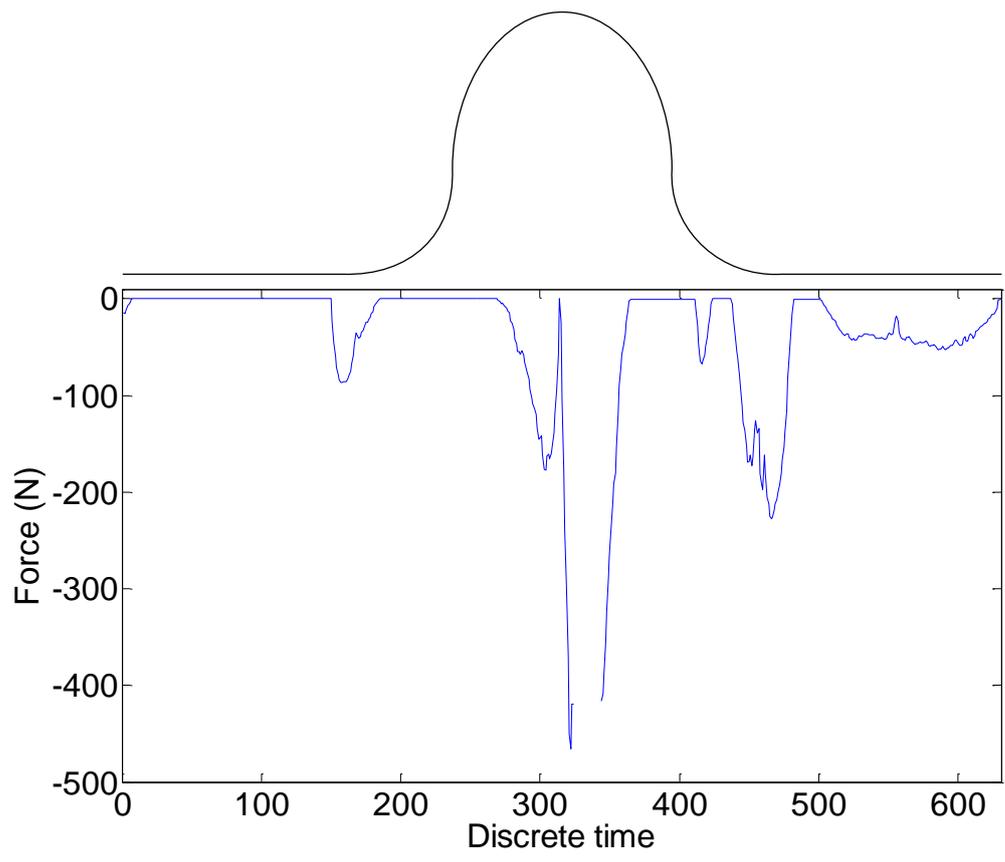

**Figure 18**

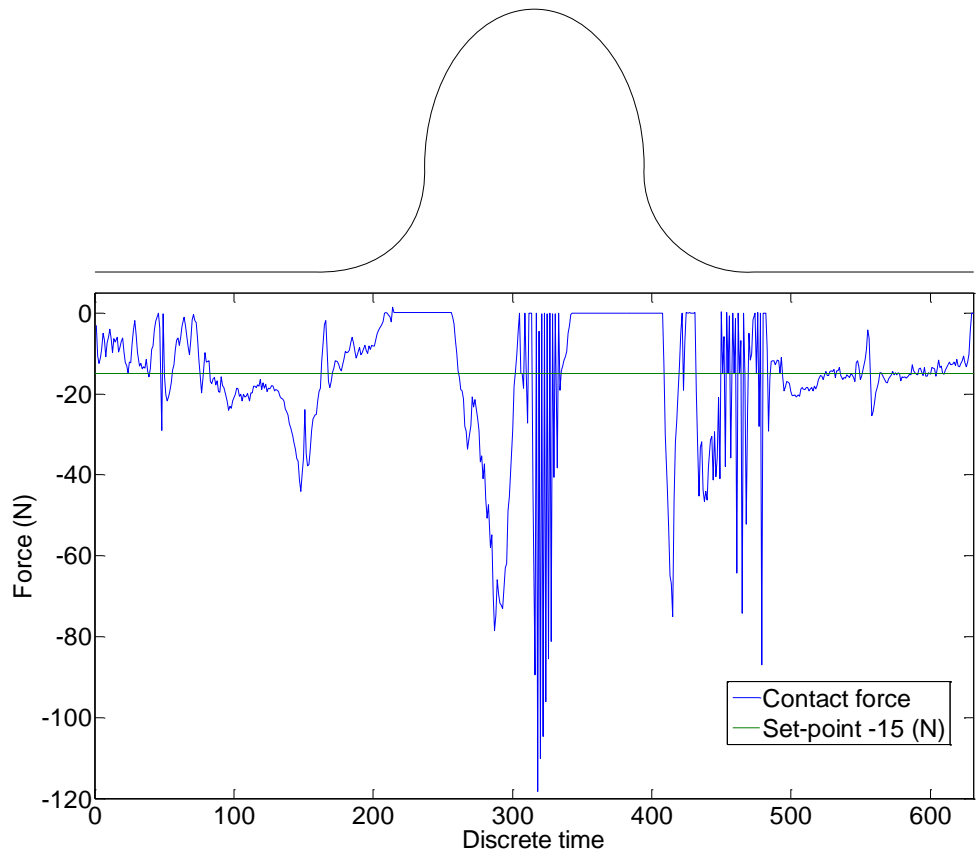